\documentclass{article} 
\usepackage{iclr2024_conference,times}

\usepackage{amsmath,amsfonts,bm}

\def\eqref#1{equation~\ref{#1}}

\def\1{\bm{1}}

\DeclareMathAlphabet{\mathsfit}{\encodingdefault}{\sfdefault}{m}{sl}
\SetMathAlphabet{\mathsfit}{bold}{\encodingdefault}{\sfdefault}{bx}{n}

\usepackage{hyperref}
\usepackage{url}
\usepackage{cleveref}
\usepackage{booktabs}
\usepackage{multicol}
\usepackage{multirow}
\usepackage{graphicx}
\usepackage{enumitem}
\usepackage{subcaption}
\usepackage{siunitx}

\title{Stable Continual Reinforcement Learning via Diffusion-based Trajectory Replay}

{\centering
\author{Feng Chen$^{1,2}$, Fuguang Han$^{1}$, Cong Guan$^{1}$, Lei Yuan$^{1,2}$, \textbf{Zhilong Zhang}$^{1,2}$,\\ \textbf{Yang Yu}$^{1,2}$, \textbf{Zongzhang Zhang}$^{1}$\thanks{corresponding author} \\
$^1$ National Key Laboratory for Novel Software Technology, Nanjing University, China \\
\ \ \ School of Artificial Intelligence, Nanjing University, China \\
$^2$ Polixir Technologies\\
\texttt{\{chenf,hanfg,guanc,yuanl,zhangzl\}@lamda.nju.edu.cn}, \\ \texttt{\{yuy,zzzhang\}@nju.edu.cn}
}}

\newcommand{\state}{\mathcal{S}}
\newcommand{\action}{\mathcal{A}}
\newcommand{\reward}{\mathcal{R}}
\newcommand{\trans}{\mathcal{P}}
\newcommand{\expect}{\mathbb{E}}
\newcommand{\task}{\mathcal{M}}
\newcommand{\normal}[2]{\mathcal{N}({#1}, {#2})}

\newcommand{\name}{{DISTR}}

\iclrfinalcopy 
\begin{document}

\maketitle

\begin{abstract}
Given the inherent non-stationarity prevalent in real-world applications, continual Reinforcement Learning (RL) aims to equip the agent with the capability to address a series of sequentially presented decision-making tasks. Within this problem setting, a pivotal challenge revolves around \textit{catastrophic forgetting} issue, wherein the agent is prone to effortlessly erode the decisional knowledge associated with past encountered tasks when learning the new one. In recent progresses, the \textit{generative replay} methods have showcased substantial potential by employing generative models to replay data distribution of past tasks. Compared to storing the data from past tasks directly, this category of methods circumvents the growing storage overhead and possible data privacy concerns. However, constrained by the expressive capacity of generative models, existing \textit{generative replay} methods face challenges in faithfully reconstructing the data distribution of past tasks, particularly in scenarios with a myriad of tasks or high-dimensional data. Inspired by the success of diffusion models in various generative tasks, this paper introduces a novel continual RL algorithm DISTR (Diffusion-based Trajectory Replay) that employs a diffusion model to memorize the high-return trajectory distribution of each encountered task and wakeups these distributions during the policy learning on new tasks. Besides, considering the impracticality of replaying all past data each time, a prioritization mechanism is proposed to prioritize the trajectory replay of pivotal tasks in our method. Empirical experiments on the popular continual RL benchmark \texttt{Continual World} demonstrate that our proposed method obtains a favorable balance between \textit{stability} and \textit{plasticity}\footnote{Here, \textit{stability} means the ability to not forget decisional knowledge from old tasks, while \textit{plasticity} denotes the capability to address subsequent tasks effectively. More details are introduced in~\Cref{sec:crl}.}, surpassing various existing continual RL baselines in average success rate. 
\end{abstract}

\section{Introduction}

Real-world decision-making applications usually face open-ended changes in scenarios, necessitating the agent to learn and adapt its policy in a life-long manner~\citep{parisi2019continual}. To serve this purpose, continual Reinforcement Learning (RL) emerges with the goal of equipping the agent with continual learning capability~\citep{khetarpal2022towards,wolczyk2021continual}. In this setting, it is typically assumed that the agent is encountered with different decision-making tasks sequentially and is not allowed to interact with the previous tasks
when learning the new one~\citep{khetarpal2022towards}. This problem setup presents a formidable challenge known as the \textit{catastrophic forgetting}~\citep{mccloskey1989catastrophic} issue. Namely, the agent is prone to forget the decisional knowledge related to previous tasks when engaging in the learning of new ones.

To address this forgetting issue, some regularization-based methods~\citep{kirkpatrick2017overcoming,aljundi2018memory} introduce an additional term outside the loss function to restrict changes in network parameters. Yet, understanding the relationship between network parameters and the final policy output is challenging, and such regularization at the parameter level may negatively impact the learning of subsequent tasks~\citep{ahn2019uncertainty}. Differently, parameter isolation methods assign distinct parameters to each task, effectively mitigating the \textit{catastrophic forgetting} issue~\citep{wang2023comprehensive}. However, these methods typically involve establishing new network branches dynamically for new tasks~\citep{rusu2016progressive} or allocating a static network architecture to continually encountered tasks~\citep{mallya2018packnet}. The former results in unrestricted architecture size, while the latter limits the network plasticity when facing the succeeding tasks~\citep{de2021continual}.
Diverging from these two methodological classifications, replay-based methods distinctly leverage an experience replay to archive data from previous tasks~\citep{wang2023comprehensive}. Despite the simplicity and efficacy inherent in this approach, its feasibility can be compromised, as it entails a growing demand for storage as the task count advances. Moreover, in certain scenarios, it may even be deemed impossible, constrained by factors such as privacy concerns that prohibit the retention of data from prior tasks~\citep{binjubeir2019comprehensive}.
To overcome these drawbacks, some subsequent approaches suggest the utilization of generative models to produce pseudo rehearsals instead of storing the original data directly, known as \textit{generative replay} methods.

More specifically, \textit{generative replay} methods involve training generative models, such as VAE~\citep{kingma2013auto} and GAN~\citep{goodfellow2020generative}, to memorize the data distribution of the ongoing task. When new tasks arise, the generative model is used to replay prior task distributions. This approach inherits the benefits of replay-based methods while mitigating the drawbacks of directly storing data. It also emulates memory processes in biological brains~\citep{aljundi2019continual}. \textit{However, are these methods flawless?} The answer is no. Due to the limited capacity of generative models, existing \textit{generative replay} methods struggle to represent trajectory distributions, often focusing only on generating individual state data. This can lead to model collapse at specific time steps and fail to capture full trajectory distributions. Additionally, they face challenges with high-dimensional data~\citep{daniels2022model} and are generally designed for a small number of tasks~\citep{atkinson2021pseudo}.
Inspired by the success of diffusion models in various generative tasks~\citep{lugmayr2022repaint,luo2021diffusion,croitoru2023diffusion,li2022diffusion}, we explore whether these models can improve generative replay methods.
In this paper, we propose \textbf{DISTR} (abbreviated for \textbf{Di}ffu\textbf{s}ion-based \textbf{T}rajectory \textbf{R}eplay), a stable continual RL algorithm. The framework includes two modules: \textit{the decision policy} and \textit{the diffusion model}. \textit{The decision policy} learns to solve new tasks, while \textit{the diffusion model} memorizes skilled trajectory\footnote{We term trajectories sampled near the end of the policy learning process as \textit{skilled trajectories}, typically associated with high returns.} distributions. During new task learning, \textit{the diffusion model} replays old task distributions to prevent catastrophic forgetting. Given the potential infinity of tasks in continual RL, replaying data from all previous tasks is impractical. To address this, we introduce a prioritization mechanism that selectively replays skilled trajectories of key tasks, enhancing the approach's scalability for real-world scenarios.

We empirically evaluate the effectiveness of our \name~on the prevalent continual RL benchmark \texttt{Continual World}~\citep{wolczyk2021continual}.
The experimental results indicate that our approach surpasses various categories of continual RL baselines in terms of the average success rate.
Evaluation results on different evaluation metrics demonstrate that \name~achieves a favorable trade-off between stability and plasticity, resulting in overall better continual learning capability. Besides, visualization results show the diffusion model can reconstruct the trajectory distribution well. 

\section{Background}
\subsection{Continual Reinforcement Learning}
\label{sec:crl}

Continual Reinforcement Learning aims to train the agent on a set of decision-making tasks sequentially, where each task can have specific definitions and the agent is not allowed to access the previous tasks during this sequential learning paradigm~\citep{khetarpal2022towards}.

In specific, each task in the domain of Reinforcement Learning (RL) can be formalized as a Markov Decision Process (MDP) $\mathcal{M}:=\langle \state, \action, \trans, \reward, \gamma\rangle$, where $\state, \action$, and $\gamma$ respectively indicate the state space, action space and discount factor of the task. At each timestep $t$, the agent selects an action $a\in\action$, which leads to the environment transferring into the next state $s'\in\state$ according to the transition function $\trans (s'|s,a)$. At the same time, a reward $r=\reward(s,a)$ is returned to the agent as the feedback of the executed action. The learning objective for one individual task is to maximize the discounted return $\expect_{s_{0:\infty},a_{0:\infty}}\left[\sum_{t=0}^\infty \gamma^t \reward(s_t,a_t)|s_0=s,a_0=a\right]$. In practice, we can employ the discounted return to evaluate the agent's policy for an individual task or use other specific metrics, such as the success rate.

When it comes to the continual RL, we can define the whole learning problem as a Continual MDP (CMDP) $\mathcal{C}$, which is constituted with a set of decision-making tasks, i.e., $\mathcal{C} = \{ \mathcal{M}^{(z)}|z\in \mathbb{Z}_{+}\}$, where $\mathbb{Z}_{+}$ denotes the set of non-negative integers. Each task $\mathcal{M}^{(z)}$ is defined by a tuple $\langle \state^{(z)}, \action^{(z)}, \trans^{(z)}, \reward^{(z)}, \gamma^{(z)}\rangle $ as described before, and all tasks appear in a specific sequence. When the agent is engaging in the $k$-th task $\task^{(k)}$, it is typically assumed that the agent can not access the previous tasks $\{\task^{(i)}|i\leq k\}$ any more. In this problem setup, the agent is expected to perform well across all encountered tasks, embodying the essence of lifelong learning. Thus, the average performance across all these tasks often serves as the metric in this problem setting. 

A classic challenge in continual (reinforcement) learning is the \textit{stability}-\textit{plasticity} dilemma. \textit{Stability}, in this context, refers to the ability of the agent to maintain satisfactory performance on previous tasks when confronted with subsequent tasks, avoiding the forgetting of knowledge from old tasks. \textit{Plasticity}, on the other hand, indicates that the agent policy retains sufficient expressive capability for learning new tasks, even with positive forward transfer.

\subsection{Diffusion Models}


Initially proposed by~\citet{sohl2015deep}, diffusion models mimic the thermodynamic diffusion process to model the generative process of new data. With the proposal of the framework Denoising Diffusion Probabilistic Model (DDPM)~\citep{DDPM}, recently diffusion models showcased superior generative capabilities than prior generative models such as Variational Autoencoders (VAE)~\citep{kingma2013auto} and Generative Adversarial Networks (GAN)~\citep{goodfellow2020generative}, achieving remarkable success in various domains~\citep{lugmayr2022repaint,luo2021diffusion,croitoru2023diffusion,li2022diffusion}. 

The core idea of this framework is a parameterized multi-step denoising process to generate real data $x^0$ from random noise $x^T\sim \normal{0}{I}$ ($I$ stands for the identity matrix), which can be seen as the reversed process of the forward diffusion process $x^{0:T}$. Specifically, each forward transition $q(x^t|x^{t-1})$ can be seen as adding a Gaussian noise with scheduled variance $\beta^t$, which indicates that:
\begin{equation}
    x^t = \sqrt{\alpha^t} x^{t-1} + \sqrt{1-\alpha^t} \epsilon^t,
\end{equation}
where $\alpha^t = 1 - \beta^t$, $\epsilon^t\sim \normal{0}{I}$. This results in $x^t$ being a sample from Gaussian distribution $\normal{\sqrt{\alpha^t}x^{t-1}}{\sqrt{1 - \alpha^t}}$. Generally, it is required that $\lim_{T\rightarrow\infty}\alpha^T = 0$ to ensure that when $T$ is sufficiently large $x^T$ is a random noise from standard Gaussian distribution. Then, to generate real data from random noise $x^T\sim\normal{0}{I}$, a denoising network is applied to parameterize the denoising process $p_\theta(x^{t-1}|x^t)$, which is reversed process of the forward transition. More specifically, a parameterized network $\epsilon_\theta(x^t, t)$ is applied to predict the added noise $\epsilon^t$, with which we can express the mean of variable $x^{t-1}$ by:
\begin{equation}
    \mu_\theta(x^t, t) = \frac{1}{\sqrt{\alpha_t}}\left(x_t - \frac{\beta_t}{\sqrt{1 - \bar{\alpha}_t}}\epsilon_\theta(x^t,t)\right),
\end{equation}
where $\bar{\alpha}_t=\prod_{i=0}^t\alpha_i$. This equation allows us to obtain real data $x^0$ through denoising the random noise step by step, utilizing the parameterized noise prediction network $\epsilon_\theta$. The denoising network $\epsilon_\theta$ here is optimized through minimizing the loss $L(\theta) = \expect_{x^0, t, \epsilon}\left[\|\epsilon - \epsilon_\theta(x^t,t)\|\right]$, where the symbolic denotes the $L_1$ loss, as used in our method.
\begin{figure}[t!]
    \centering
    \includegraphics[width=\linewidth]{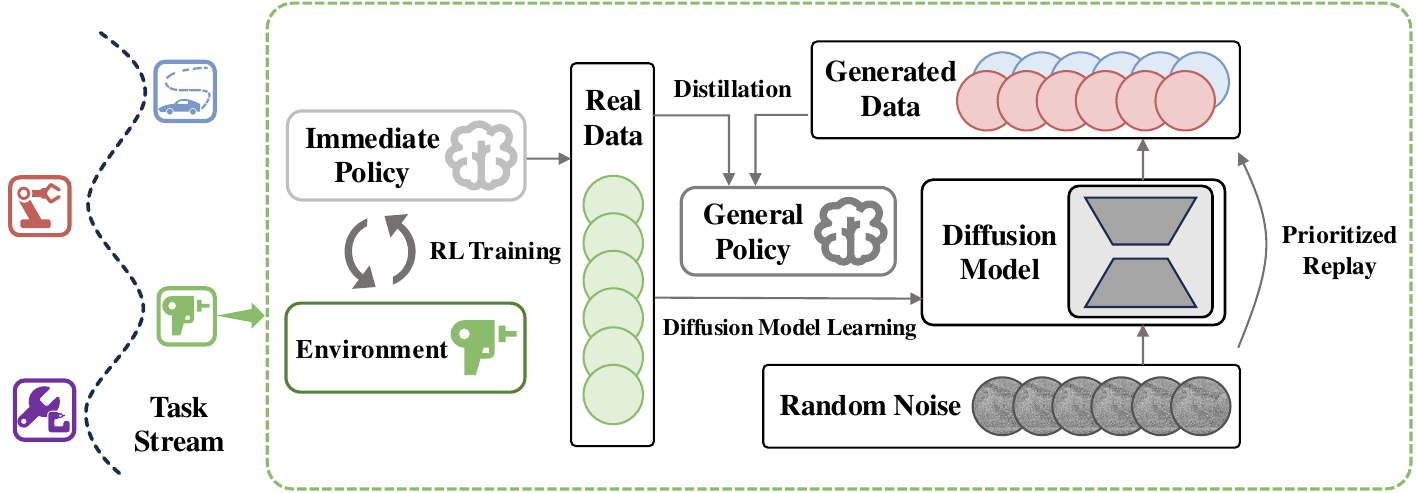}
    \caption{Overall framework of our proposed DISTR.}
    \label{fig:framework}
\end{figure}
\section{Method}
\newcommand{\loss}{\mathcal{L}}
\newcommand{\dataset}{\mathcal{D}}

Our proposed method \name~involves the joint learning of both a decision policy and a diffusion model, where the decision policy learns how to tackle the ongoing decision tasks, while the diffusion model is employed to replay the data distribution of prior tasks, mitigating the forgetting issue of the decision policy. The whole framework of \name~is depicted in~\Cref{fig:framework}. 

In this section, we introduce the details of our proposed DISTR. To begin with, we provide the training details of the decision policy, where we decouple it into a short-term \textit{immediate policy} and a long-term \textit{general policy}, ensuring each policy optimized with one specific type of learning objective to stabilize the learning process. Then we proceed to illustrate how the diffusion model is trained to continually fit the data distribution of each coming task. Lastly, a prioritization mechanism is proposed to prioritize data replay, relieving the concerns of encountering infinite tasks. 

\subsection{Decision Policy Training}
\label{sec:policy}
Upon the arrival of each new task, the agent policy is typically trained utilizing RL algorithms, e.g., SAC~\citep{haarnoja2018soft}, to learn the decisional knowledge of the present task. These off-the-shelf RL algorithms have proved their efficacy in various decision-making domains~\citep{wang2020deep}.
However, excessively updating the RL loss for the current task may induce the agent policy to forget decisional knowledge of past ones. 
To address this issue, a straightforward approach is to introduce extra Behavior Cloning (BC) terms, leveraging skilled trajectories of past tasks if available, apart from the main RL loss term.
This indicates that when learning on the $k$-th task, we actually optimize the agent policy with the following objective:
\begin{equation}
    \begin{aligned}
    \min_\phi \loss^{(k)}(\phi) &= \loss^{(k)}_{\rm{SAC}}(\phi) + \sum_{i=0}^{k-1} \loss_{\rm{BC}}^{(i)}(\phi)\\
    \loss_{\rm{BC}}^{(i)}(\phi) &= \expect_{\tau\in\dataset^{(i)}}\expect_{(s,a)\in\tau} \left[-\log \pi_\phi (a|s)\right],
    \end{aligned}
\end{equation}
where $\phi$ denotes the parameters of agent policy while $\dataset^{(i)}$ indicates the skilled trajectories of task $i$ ($i < k$). This learning objective regularizes the agent policy to retain the decisional knowledge of past tasks when learning the new one. Besides, accounting for the previously discussed issues caused by directly storing data from past tasks, we utilize a diffusion model to generate the trajectories instead of storing the real data. More details concerning the diffusion model training will be introduced in the next section.
While it is a simpler approach to update the agent policy through optimizing this objective function directly, we emphasize that it may bring about some specific drawbacks. For one thing, the scales of the introduced BC regularization loss terms typically differ from the main RL loss, thus necessitating a dedicate coefficient tuning to balance them. For another, multiple loss terms of different categories may introduce gradients that conflict with each other, causing negative impact on the policy learning of the new task. 

To address the aforementioned issues, we propose a two-folded policy training scheme, decoupling the agent policy into a short-term \textit{immediate policy} $\pi_{\hat{\phi}}$ for decisional knowledge extraction of the current task and a long-term \textit{general policy} $\pi_\phi$ for general decision knowledge distillation. More specifically, when learning the $k$-th task, we first let the immediate policy inherit the parameters of the general policy from the previous round. Then, we update the immediate policy solely with the RL loss $\loss^{(k)}_{\rm{SAC}}(\hat{\phi})$, resulting in a skilled policy for the new task. Afterwards, we select top $N_{\rm{traj}}$ trajectories near the end of learning on the present task as $\dataset^{(k)}$, where $N_{\rm{traj}}$ is a hyper-parameter, and utilize the diffusion model $G_\theta$ to generate pseudo skilled trajectories of past tasks, $\{\dataset^{(i)}|i<k\}$. Then we train the general policy purely with BC losses:
\begin{equation}
    \max_{\phi} \sum_{i=0}^k\loss^{(i)}_{\rm{BC}}(\phi) =  \sum_{i=0}^k\expect_{\tau\in\dataset^{(i)}}\expect_{(s,a)\in\tau} \left[-\log \pi_{\phi}(a|s)\right].
\end{equation}
Through proposing this two-folded training scheme, we ensure that each training stage involves only one category of losses, avoiding extra coefficient tuning effort due to the scale inconsistency among loss terms. Also, this practice shields the new task learning from the influence of other optimization objectives, thereby stabilizing the learning of the new task. 

\subsection{Diffusion Model Training}
\label{sec:diffusion}
In the last section, we have introduced the training details of the agent policy, which is a two-folded training scheme involving two policy networks. In that process, we rely on the diffusion model to generate pseudo skilled trajectories of the past tasks ($\{\dataset^{(i)}|i<k\}$ when learning task $k$). In this section, we will detail how we train the diffusion model.

To equip the diffusion model in our method with the ability to memorize the decisional knowledge of tasks till now, we continually train the diffusion model on skilled trajectories datasets of different tasks. More specifically, upon finishing the learning of the immediate policy $\pi_{\hat{\phi}}$ on task $k$, we utilize the selected skilled trajectories $\dataset^{(k)}$ the same as that in~\Cref{sec:policy} to train the diffusion model $G_{\theta}$. Unlike the typical practice in previous works that trained the generative model on individual states, we directly utilize the diffusion model to fit the whole trajectory distribution, which means that our diffusion model is employed to generate a complete trajectory each time. This practice circumvents the issue of imbalanced generation of states at different trajectory timesteps and eliminate the need to condition the generative model on the trajectory timestep and iterate over timesteps for data generation as seen in~\citet{yue2023t}.

Moreover, as we train the diffusion model task-by-task, the forgetting issue may also loom over the diffusion model~\citep{zajkac2023exploring}, which means that it may overfit to the learning of the current task while forgetting the data distribution of previous tasks. To address this issue, we adopt a self-cloning technique for the diffusion model training, which means that aside from the real trajectories of the current task, the generated trajectories of past tasks are also adopted to train the diffusion model together. To put it in another way, we incorporate the generated $\{\dataset^{(i)}|i<k\}$ with the real data $\dataset^{(k)}$ on task $k$ to train the diffusion model, wherein we condition the sampling on task id to disguish the data distribution of different tasks.

\subsection{Replay Prioritization Mechanism}
\label{sec:priority}
Considering that there may exist an infinite number of tasks in the continual RL problem setting, it is impractical for DISTR to replay skilled trajectories of all past tasks each time. To mitigate this issue, a straightforward idea is to allocate the limited trajectory replay chances to more pivotal tasks. We claim that the pivotal tasks referred to here should possess the following two significant features: 1) \textit{\textbf{Vulnerability}}, indicating the tasks that are more sensitive to forgetfulness; 2) \textit{\textbf{Specificity}}, signifying tasks hard to avoid forgetting by replaying data from other tasks. The question then arises: \textit{how do we quantify these two metrics for different tasks and recognize the pivotal tasks? }

To answer this question, we propose a novel prioritization mechanism to prioritize replaying the skilled trajectories of pivotal tasks. For \textit{Vulnerability}, we quantify the robustness of performance for each task when there is a slight deviation in policy outputs. Specifically, after learning the immediate policy on task $k$, we obtain its success rate on task $k$ as $s_k$. Then, we repeatedly add random noise perturbations to the policy outputs and retest to obtain $\hat{s}_k$, the estimated success rate. We utilize $s_v^{(k)} = s_k - \hat{s}_k$ to quantify the \textit{Vulnerability}, which means that tasks with large $s_v^{(k)}$ are more \textit{Vulnerable} because they face greater performance drops when the policy output deviates. On the other hand, for \textit{Specificity}, we utilize the initial success rate $s_s^{(k)}$ on task $k$ upon its arrival to estimate the relevance of previous tasks to task $k$ in terms of decision making. It means that tasks with larger $s_s^{(k)}$ may benefit from the trajectory replay of previous tasks and are less $\textit{Specific}$. Overall, we compute the priority of each task as $(s_v^{(k)} + 1- s_s^{(k)})/2$, and replay the skilled trajectories of each task with the normalized probability.

\section{Experiments}
\label{sec:exp}

We mainly validate our proposed approach DISTR on a continual RL benchmark~\texttt{Continual World}~\citep{wolczyk2021continual}, which constructs task sequences based on the MetaWorld~\citep{yu2020meta} environment. In this section, the experimental results are provided to answer the following questions: 1) Can \name~perform better than existing continual RL methods on the provalent benchmark? (See \Cref{sec:main_performance}). 2) How well does DISTR achieve the balance between the stability and plasticity in the continual learning process? (See \Cref{sec:curve_analysis}). 3) Do diffusion models really excel in replaying data distribution of previous tasks? (See \Cref{sec:visualization}).

\paragraph{Baselines} In our experiments, we compare our method with different categories of continual RL baselines as follows:
\begin{itemize}[itemsep=0pt,parsep=0pt,topsep=0pt,partopsep=0pt]
    \item \textbf{Finetune} is the most basic method that directly trains the policy sequentially on the stream of decision tasks without any extra algorithm design. 
    \item \textbf{EWC}~\citep{kirkpatrick2017overcoming} is a regularization-based method that conducts weighted constraints on network parameters to discourage deviations from the previous network. Its importance weight is approximated using the Fisher information matrix.
    \item \textbf{MAS}~\citep{aljundi2018memory} is also a regularization-based method like EWC. But it obtains the importance weight by approximating the influence of each parameter on the policy output.
    \item \textbf{PackNet}~\citep{mallya2018packnet} serves as a parameter isolation method that allocates a fixed network architecture into different tasks through iteratively pruning, freezing and retraining the network parameters.
    \item \textbf{RePR}~\citep{atkinson2021pseudo} is a generative replay method that utilizes GAN to generate individual observations of previous tasks to mitigate forgetting.
\end{itemize}
\paragraph{Metrics} Note that we denote the success rate on task $j$ after the training of task $i$ as $s_i(j)$. Supposing that we totally encounter $N+1$ tasks, we evaluate different methods in terms of the following three different metrics:
\begin{itemize}[itemsep=0pt,parsep=0pt,topsep=0pt,partopsep=0pt]
    \item \textbf{Average Performance} The \textit{average performance} indicates the average success rate across all the $N+1$ tasks after all training, which is computed by $\frac{1}{N+1}\sum_{i=0}^N s_N(i)$.
    \item \textbf{Forward Transfer} \textit{Forward transfer} estimates the performance gain compared to a reference policy individually trained on the task. To quantify it, we obtain success rate $s^{\rm{ref}}(k)$ for each task $k$ through single task training. Then we quantify the \textit{forward transfer} as $\frac{1}{N+1}\sum_{i=0}^NFT_i$, where
    \begin{equation}
        FT_i = \frac{S_i - S^{\rm{ref}}_i}{1 - S^{\rm{ref}}_i}, \quad S^{\rm{ref}}_i= \frac{s^{\rm{ref}}(i)}{2}, \quad S_i = \frac{s_i(i) + s_{i-1}(i)}{2}.
    \end{equation}.
    \item \textbf{Forgetting} \textit{Forgetting} estimates the extent of task forgetting which is computed by $\frac{1}{N+1}\sum_{i=0}^NF_i$, where $F_i = s_i(i) - s_N(i)$.
\end{itemize}

\begin{table}[t!]
    \centering
    \scalebox{0.82}{
    \begin{tabular}{ccccccc}
        \toprule
        \multirow{2.5}{*}{Method} & \multicolumn{3}{c}{CW5} & \multicolumn{3}{c}{CW10} \\
        \cmidrule[0.5pt]{2-7}
        & Avg. Performance $\uparrow$ & F. T. $\uparrow$ & Forgetting $\downarrow$ & Avg. Performance $\uparrow$ & F. T. $\uparrow$ & Forgetting $\downarrow$ \\
        \midrule
        Finetune & {\ \ 7.0}$\pm$6.8 & {\ \ \ 4.3}$\pm${1.7\ \ \ \ \ } & 71.7$\pm${1.2\ \ } & {\ \ 7.8}$\pm$4.5 & {\ \ -2.4}$\pm${4.7\ \ \ \ \ } & 63.2$\pm${2.1\ \ } \\
        \midrule
        EWC & 35.2$\pm$3.0 & {-21.5$\pm$1.3\ \ \ \ \ } & {\ \ 2.5}$\pm${1.5\ \ } & 19.7$\pm${7.3} & -40.9$\pm${3.7\ \ \ \ \ } & {\ \ 3.0}$\pm${2.5\ \ } \\
        MAS & 26.6$\pm$6.4 & {-27.4$\pm$4.7\ \ \ \ \ } & {\ \ 4.0}$\pm${1.0\ \ } & 17.9$\pm${7.8} & -43.4$\pm${7.1\ \ \ \ \ } & {\ \ 1.8}$\pm${0.7\ \ } \\
        PackNet & 79.9$\pm$0.1 & {\ \ \ 5.9}$\pm${1.1\ \ \ \ \ } & {\ \textbf{-0.3}}$\pm${0.9\ \ } & 80.8$\pm$0.9 & {\ \ \ 6.8}$\pm${1.3\ \ \ \ \ } & {\ \textbf{-0.3}}$\pm${2.3\ \ } \\
        RePR & 34.0$\pm$1.6 & {\ \ -4.8}$\pm${3.0\ \ \ \ \ } & 27.3$\pm${2.5\ \ } & 32.3$\pm${3.1} & {\ \ -5.5}$\pm${2.2\ \ \ \ \ } & 33.3$\pm${0.9\ \ } \\
        DISTR & \textbf{84.8}$\pm$2.3 & {\ \textbf{16.2}}$\pm${1.6\ \ \ \ \ } & {\ \ 4.7}$\pm${2.0\ \ } & \textbf{81.2}$\pm$0.2 & {\ \textbf{14.7}}$\pm${3.7\ \ \ \ \ } & {\ \ 3.9}$\pm${0.8\ \ } \\
        \bottomrule
    \end{tabular}}
    \caption{Main experimental results on Continual World on three metrics, including Avg. Performance (\textit{Average Performance}), F. T. (\textit{Forward Transfer}) and Forgetting (\textit{Forgetting}). ``$\uparrow$'' means more is better, while ``$\downarrow$'' means less is better. The best results for each metric are \textbf{bolded}.}
    \label{tab:main_res}
\end{table}

\subsection{Main Results on Continual World}
\label{sec:main_performance}

We empirically evaluate all the methods on the CW5 and CW10\footnote{CW10 consists of the task sequence \textit{hammer-v2}, \textit{push-wall-v2}, \textit{faucet-close-v2}, \textit{push-back-v2}, \textit{stick-pull-v2}, \textit{handle-press-side-v2}, \textit{push-v2}, \textit{shelf-place-v2}, \textit{window-close-v2}, \textit{peg-unplug-side-v2} from Meta-World, while CW5 consists of the first five tasks from the sequence.} task sequences, wherein the results are listed in~\Cref{tab:main_res}. Specifically, we present the results on three metrics, respectively \textit{average performance}, \textit{forward transfer} and \textit{forgetting}, thus providing a comprehensive analysis of the performance of different methods.
Actually, it can be observed that our proposed DISTR obtains the highest final average success rate on both of these two task sequences, validating the overall efficacy of our approach for continual RL.

In contrast, the Fine-tune method fails to exhibit any capability for continual learning, highlighting the \textit{catastrophic forgetting} issue and calling for effective algorithm design to construct agents endowed with lifelong learning capabilities. Regularization-based methods EWC and MAS perform slightly better than Fine-tune, yet still fall short of achieving satisfactory results. Actually, these two methods can achieve relative decent anti-forgetting effect benefiting from the parameter regularization. However, the penalty imposed on parameter-level changes significantly hampers the forward transfer effect of these methods, resulting in relatively low overall performance. The method that aligns with our performance most is PackNet, which achieves nearly zero-forgetting through parameter isolation. However, due to its iterative pruning of the parameters, it results in insufficient network plasticity for subsequent tasks, leading to a slightly lower forward transfer. This underscores the primary reason why our proposed DISTR overall surpasses PackNet, achieving a $5\%$ improvement for average success rate on CW5.

In general, these results validate the superiority of our approach utilizing generative replay compared to other categories of methods. Also, we have incorporated another generative replay method RePR. The significant advantage of our approach over RePR confirms the contribution of employing diffusion models and underscores the overall effectiveness of our methodological framework.

\subsection{Analysis of the Continual Learning Process}
\label{sec:curve_analysis}

\begin{figure}[htbp]
  \centering
  \includegraphics[width=\linewidth]{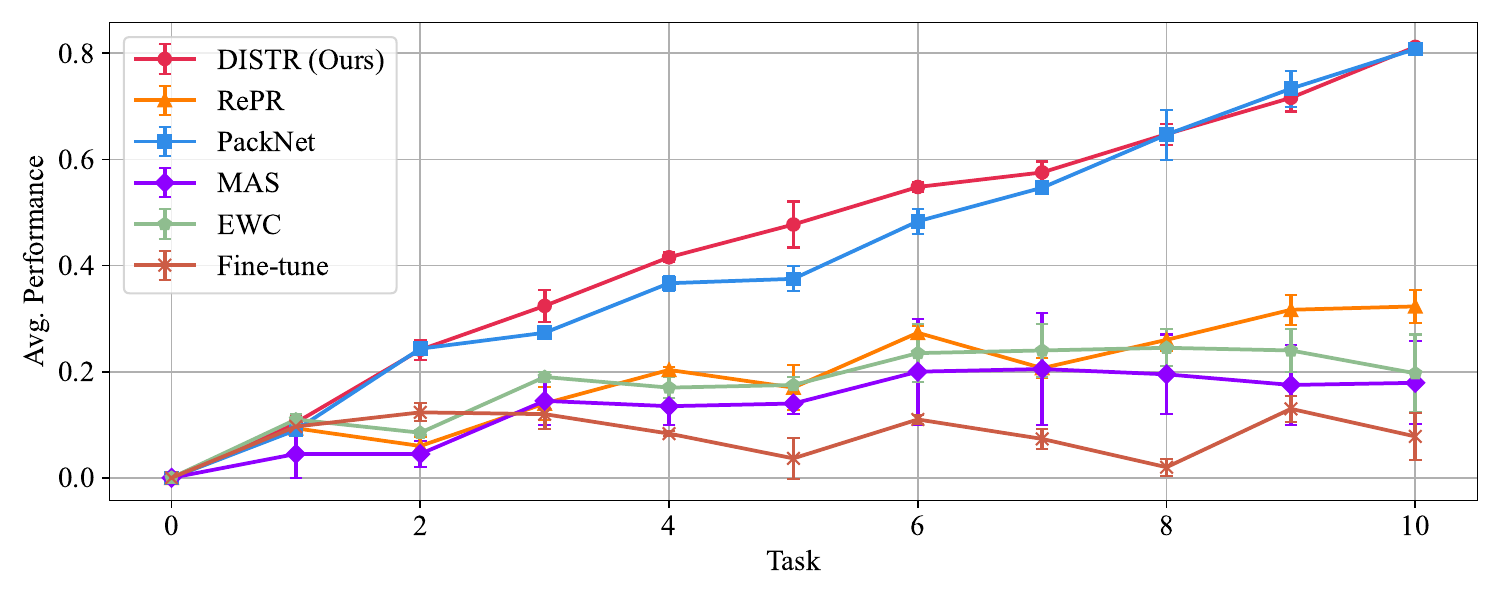}
  \caption{Average success rate on CW10 of different methods during the continual learning.}
  \label{fig:main_curve}
\end{figure}

In~\Cref{sec:main_performance}, we provide the final performance of different methods on two task sequences.
Herein, for a more comprehensive understanding of the learning details, we depict the learning curve of various methods on CW10 in~\Cref{fig:main_curve}. This figure illustrates the fluctuations in average success rate across ten tasks throughout the continual learning process.

Specifically, from this figure, we can analyse the performance in terms of stability and plasticity of different methods. For example, the average performance of Fine-tune consistently remains within $0.15$ throughout the entire process, which indicates that Fine-tune consistently fails to surpass performance on more than two tasks, revealing its terrible forgetting issue. 
For regularization-based methods, it is noticeable that their curves fail to ascend at many points. For example, MAS stays similar performance from tasks $3$ to $5$, indicating that MAS fails to solve tasks $3$ and $4$.
This phenomenon reveals that the parameter-level regularization hinders these methods from successfully learning effective policies for many subsequent tasks, indicating low plasticity. Despite its overall improved performance, PackNet, similar to these methods, exhibits difficulty in learning task $4$. In contrast, our proposed DISTR exhibits relatively better average success rate performance and stable learning process, revealing that it achieves a qualified trade-off between stability and plasticity.

\begin{figure}[t!]
    \centering
    \begin{subfigure}[b]{0.32\linewidth}
        \includegraphics[width=\textwidth]{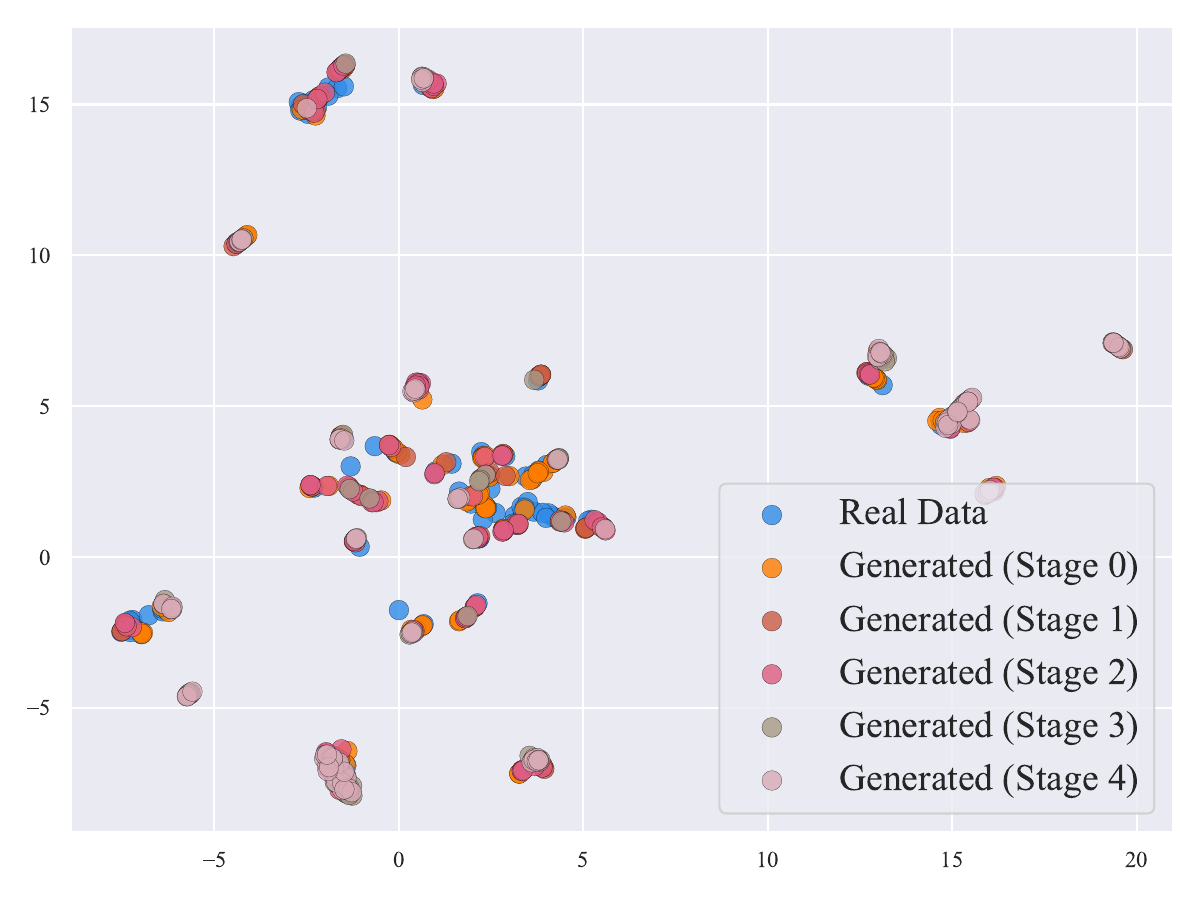}
        \caption{}
        \label{fig:scatter1}
    \end{subfigure}
    \hfill
    \begin{subfigure}[b]{0.32\linewidth}
        \includegraphics[width=\textwidth]{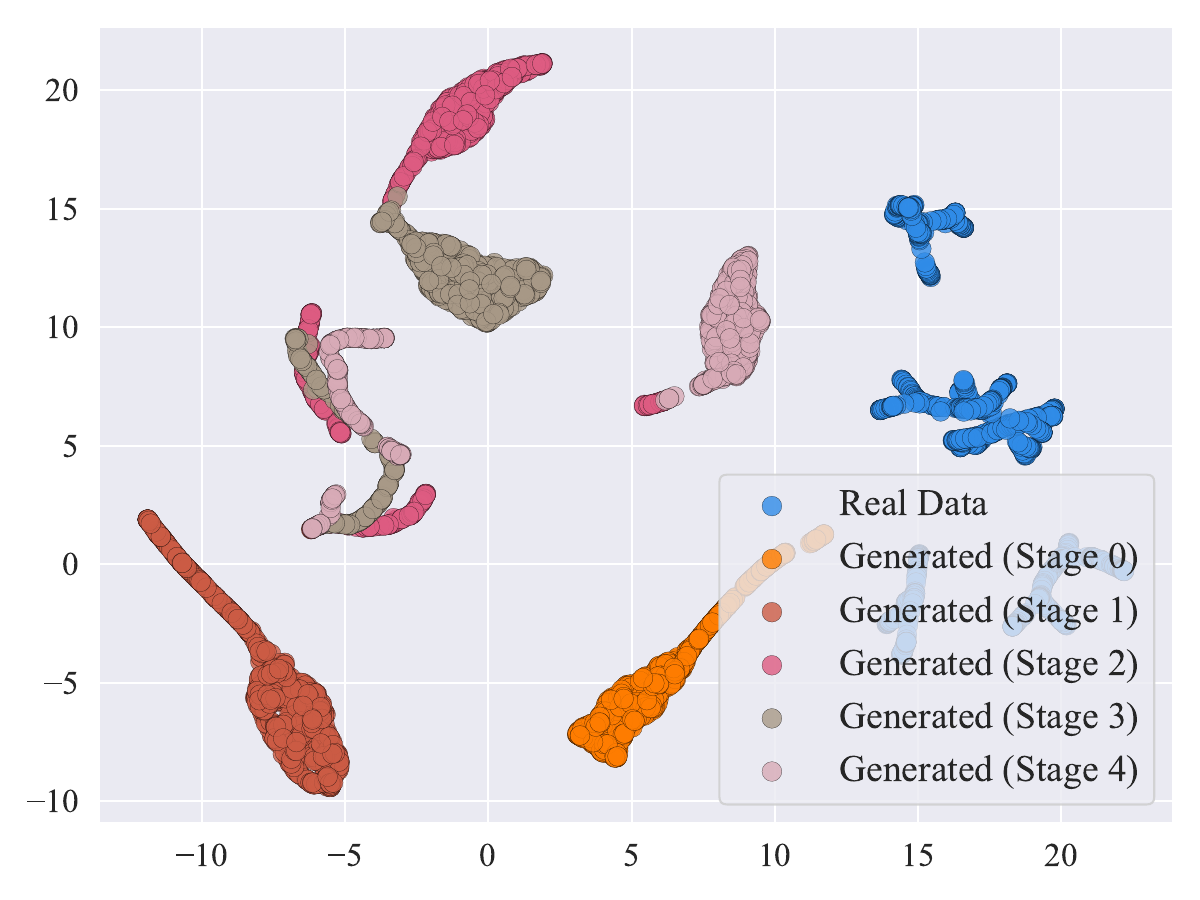}
        \caption{}
        \label{fig:scatter2}
    \end{subfigure}
    \hfill
    \begin{subfigure}[b]{0.32\linewidth}
        \includegraphics[width=\textwidth]{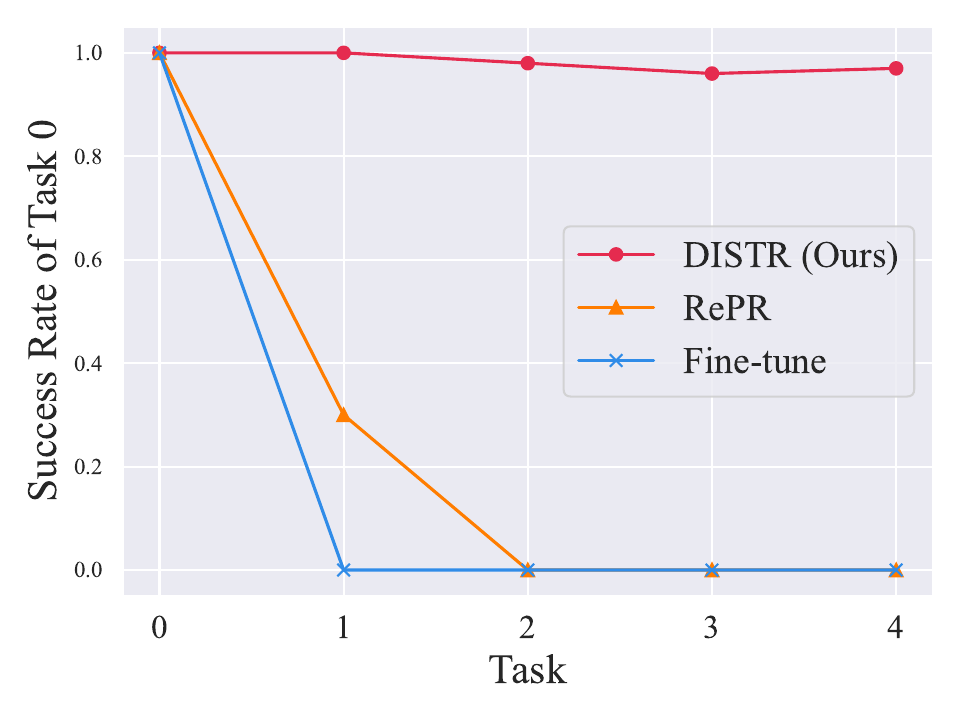}
        \caption{}
        \label{fig:scatter3}
    \end{subfigure}
    \caption{These figures illustrate the generative replay visualization study. Figure (a) depicts the visualization result of our proposed DISTR, while Figure (b) depicts that of RePR. In these figures, the generated data is produced utilizing the generative models at the end of training for task $0$-$4$. Figure (c) illustrates the performance changes on task $0$ of different methods when ending the training of task $0$-$4$.}
    \label{fig:trajectory_visualization}
    \vspace{-1.5em}
\end{figure}

\subsection{Visualization Analysis of Generative Replay}
\label{sec:visualization}

While the previous experimental results have showcased the superiority of our proposed DISTR over other continual RL baselines on the widely used benchmark \texttt{Continual World}, it is still unclear about the efficacy of the diffusion model we employed in memorizing the trajectory distribution. In this section, we answer this question via conducting visualization analysis for the generative replay. Specifically, we utilize the diffusion model trained after each task in CW5 to generate trajectory data for task $0$. Subsequently, we perform UMAP~\citep{mcinnes2018umap} dimensionality reduction on these generated trajectories together with the real data. The results are visualized as scatters in~\Cref{fig:scatter1}.
We can observe that the generated trajectories cover the real data well, indicating that our diffusion model overall faithfully reconstructs the original trajectory distribution. For comparison, we also provide the visualization results for RePR that utilizes GAN for generative replay in~\Cref{fig:scatter2}. The results show that the generated data does not cover the real data distribution well, which answers the large performance drop of RePR on task $0$ after the training of task $2$. However, our DISTR retains high performance on task $0$, indicating the efficacy of our generative replay.

\section{Related Work}

\subsection{Continual Learning for Decision Making}

In real world applications, assuming a static problem scenario is usually impractical, necessitating the research about continual learning methods~\citep{parisi2019continual,wang2023comprehensive}. Currently, there mainly exist three categories of mainstream approaches, respectively regularization-based methods, parameter isolation methods and replay-based methods~\citep{de2021continual}. Despite not originally designed for continual RL setting which involves a stream of decision-making tasks, most of these methods can be adapted for continual RL. For example,~\citet{wolczyk2021continual} built a continual RL benchmark and tested various existing categories of continual learning methods.

Among these methods, replay-based approaches not only involve methods that directly store data~\citep{rebuffi2017icarl,chaudhry2019tiny} but also incorporate methods that utilize generative replay for concerns of storage overhead or data privacy~\citep{wang2023comprehensive}. RePR~\citep{atkinson2021pseudo} employed GAN to replay the raw observations of past tasks, and regularized the policy output on these observations to not drift a lot. However, due to the expressive capability limitation of GAN, it only works well on short sequence of tasks ($3$ in original paper). Considering that it is hard to replay the raw features accurately, ~\citet{daniels2022model} proposed to replay the hidden representations with VAE, which is more applicable in some scenarios. Nevertheless, it requires an extra random replay buffer storing raw data to mitigate representation shift issue, which conflicts with the original intent of \textit{generative replay} methods. For continual RL, there also exist some works applying model-based generative replay~\citep{gao2021cril}, which have faced criticism due to the compounding error issue of the world model.

Different from these methods, our method \name~directly generates the state-action trajectories benefiting from the powerful expressive capacity of diffusion model. It does not require any extra replay buffer and empirically works well in settings of long task sequence. Besides, its model-free paradigm frees it from the compounding error issue when generating trajectories. The most close to ours are two concurrent works, t-DGR~\citep{yue2023t} and CuGRO~\citep{liu2024continual}, which also applied diffusion models to facilitate continual decision-making. t-DGR considered a continual imitation learning setting, requiring expert datasets for all tasks, which is not always applicable in real-world, and it still re-generated individual states. Similarly, CuGRO also utilized diffusion models to replay previous states, while it innovatively utilized diffusion models as both state generators and policies, with a focus on offline continual settings. DISTR differs by directly replaying entire trajectories to address online continual RL problems and proposing a prioritization mechanism to enhance performance. Further comparison with these two works in future is highlighted.

\subsection{Diffusion Models in Reinforcement Learning}

Diffusion models as a class of generative models recently have showcased impressive generative performance across various domains~\citep{lugmayr2022repaint,luo2021diffusion,croitoru2023diffusion,li2022diffusion}, especially in the task of text-to-image~\citep{text2image1,text2image2}, surpassing previous generative models in generation quality and training stability. 
Considering the powerful generative capability of diffusion models, more and more researchers started to employ diffusion models to solve problems in the domain of RL. 

One type of related works adopted diffusion models as planners~\citep{janner2022planning,liang2023adaptdiffuser,ni2023metadiffuser,zhu2023madiff}, which has been claimed to avoid suffering from compounding error issue~\citep{janner2022planning} due to the non-regressive planning scheme compared to the traditional model-based methods~\citep{moerland2023model,luo2022survey}. In some cases, guided-sampling techniques were adopted to generate trajectories with higher returns or satisfying specific constraints~\citep{liang2023adaptdiffuser,xiao2023safediffuser}. Besides, there also exist research works employing diffusion models as the policy directly~\citep{zhu2023diffusion}. The powerful expressive capabilities for multi-model distributions help diffusion-based policies obtain superior performance in both offline RL~\citep{wang2022diffusion} and imitation learning~\citep{pearce2022imitating,reuss2023goal} settings.

Apart from these two types of related works, one another category of methods utilized diffusion models as data synthesizers~\citep{zhu2023diffusion,zhang2023flow,lu2023synthetic,he2023diffusion}.~\citet{lu2023synthetic} trained the diffusion model on existing replay buffer and then utilized it to generate more training samples for policy learning.~\citet{he2023diffusion} employed diffusion models for data augmentation in the domain of multi-task learning, exhibiting better performance compared to the existing solutions.~\citet{zhang2023flow} utilized diffusion models to produce better trajectories leveraging the pairwise preference relationship. Our work differs from them by conducting data augmentation in continual RL, which is a new endeavor in this domain.
\section{Conclusion}

In this paper, we propose DISTR, a new continual RL method that attempts to address the Catastrophic Forgetting issue via utilizing the diffusion model to replay the expert distribution of previous tasks. A prioritization mechanism is also proposed to help prioritize the data replay, further enhancing the scalability and efficiency of our approach. On the popular continual RL benchmark \texttt{Continual World}, DISTR showcases superior performance than all existing continual RL methods, and achieves the highest average success rate across all tasks. These results demonstrate the potential of diffusion model to advance the field of continual RL, and we hope it can boost more interesting future research regarding this direction. In the future, we plan to investigate the effect of employing more than one diffusion model and whether it is possible to further advance the performance of DISTR via ensembling multiple diffusion models.

\bibliography{iclr2024_conference}
\bibliographystyle{iclr2024_conference}


\end{document}